# QuakeBERT: Accurate Classification of Social Media Texts for Rapid Earthquake Impact Assessment

Jin Han[1], Zhe Zheng[1], Xin-Zheng Lu[1], Ke-Yin Chen[1], Jia-Rui Lin[1,*]

*Corresponding author, E-mail: lin611@tsinghua.edu.cn; jiarui_lin@foxmail.com

(1. Department of Civil Engineering, Tsinghua University, Beijing, 100084, China)

**Abstract:**

Social media aids disaster response but suffers from noise, hindering accurate impact assessment and decision making for resilient cities, which few studies considered. To address the problem, this study proposes the first domain-specific LLM model and an integrated method for rapid earthquake impact assessment. First, a few categories are introduced to classify and filter microblogs considering their relationship to the physical and social impacts of earthquakes, and a dataset comprising 7282 earthquake-related microblogs from twenty earthquakes in different locations is developed as well. Then, with a systematic analysis of various influential factors, QuakeBERT, a domain-specific large language model (LLM), is developed and fine-tuned for accurate classification and filtering of microblogs. Meanwhile, an integrated method integrating public opinion trend analysis, sentiment analysis, and keyword-based physical impact quantification is introduced to assess both the physical and social impacts of earthquakes based on social media texts. Experiments show that data diversity and data volume dominate the performance of QuakeBERT and increase the macro average F1 score by 27%, while the best classification model QuakeBERT outperforms the CNN- or RNN-based models by improving the macro average F1 score from 60.87% to 84.33%. Finally, the proposed approach is applied to assess two earthquakes with the same magnitude and focal depth. Results show that the proposed approach can effectively enhance the impact assessment process by accurate detection of noisy microblogs, which enables effective post-disaster emergency responses to create more resilient cities.

# 1 Introduction

Earthquakes are potential threats to modern cities with large and concentrated populations and complex interdependent infrastructure systems (Lu et al., 2020). Extreme earthquakes (e.g., the 2008 Wenchuan earthquake, the 2010 Haiti earthquake, the 2011 Tōhoku earthquake, and the 2014 Ludian earthquake) will inflict severe damage and lead to substantial harm, extensive casualties, and significant losses (Cui et al., 2008; DesRoches et al., 2011; Mori et al., 2011; Hong et al., 2016). The sooner people are rescued from damaged areas after earthquakes, the fewer casualties will be (Ahadzadeh & Malek, 2021). So, time-sensitive responses such as emergency rescue operations should be taken to help people locate available resources, deliver assistance, and so on (Papadopoulos et al., 2023). Note that the governments' time-sensitive responses rely on timely and accurate information about the earthquake situations to make effective response decisions and improve management strategies (Yu et al., 2019). The most well-known disaster-related information acquiring methods include remote sensing and post-disaster surveying (Ahadzadeh & Malek, 2021). However, the remote sensing data has spatial, temporal, and spectral constraints (Ahadzadeh & Malek, 2021), and the post-disaster survey takes time and involves a huge quantity of labor and resources (Lowande et al., 2023). With the rapid development of the mobile internet and social platforms for sharing short and real-time information, valuable situational information for natural hazards can be disseminated across the world in real-time, which becomes more and more valuable for the development of resilient and sustainable cities (Yao et al., 2021). The contents generated by the users could be utilized in decision-support systems to help governments and citizens manage disaster risks and enable effective post-disaster emergency responses in terms of vulnerability assessment, early warning, monitoring, and evaluation (Adegoke et al., 2023). Besides, by analyzing social media texts, it is possible to effectively assess both the physical and social impacts of earthquakes and assist in the rapid recovery and reconstruction of cities.

Twitter is one of the most popular social media platforms that caught much attention of researchers (Yao et al., 2021). Weibo is the most popular social media platform in China (Chen et al., 2020). Most of the microblogs on social media platforms like Twitter and Weibo are textual. Therefore, useful information related to an earthquake, such as resources, damage, donation, or aid, should be extracted from textual microblogs (Yu et al., 2019). However, the contents of the massive microblogs on social media platforms are tanglesome, which is the main obstacle to finding insights from microblogs (Kundu et al., 2018). Therefore, natural language processing (NLP) techniques to classify microblogs into different themes should be developed to help us leverage the useful information from textual content from social media (Tounsi et al, 2023). To date, several studies have explored the classification categories and methods of microblogs. For example, informative microblogs are classified into five categories, including caution and advice, casualties and damage, donations of money, goods or services, people missing or found, and information sources (Imran et al., 2013). Based on the categories proposed by Imran, Yu added a category called Infrastructure and Resources to their classification schema (Yu et al., 2019). To extract the seismic intensity from microblogs, the microblogs are classified into four seismic intensity levels based on the seismic keyword list (Yao et al., 2021). Similar to Yao et al., Li et al. (Li et al., 2021) also classified the microblogs from tweets into four levels of damage (i.e., no damage or injury, slight damage, moderate damage with the possibility of injuries, and severe damage with the possibility of fatalities) based on long short-term memory (LSTM) model. Besides, Lv et al. devided earthquake-related microblogs into five levels (i.e., Weak earthquake, Felt earthquake, Moderate earthquake, Strong earthquake, Disastrous earthquake) based on the China Seismic Intensity Scale (Lv et al., 2023).

However, the social media data contain a large amount of noisy information which has the keywords of the earthquake but is irrelevant to the real physical loss of the target earthquake (Kundu et al., 2018; Lin et al., 2022). For example, there are also some microblogs introducing self-rescue knowledge during earthquakes, which contains keywords related to physical damage but cannot reflect the actual damage of the earthquake. A typical example of this type is "# Guizhou Bijie 4.4 magnitude earthquake # Earthquake self-rescue knowledge that must be remembered: (1) don't panic when an earthquake occurs, and (2) if the casualty needs a splint to mend the fractured limb, you can use plastic bags and cardboard to make an emergency splint bandage". This microblog contains the keyword reflecting severe damage (i.e., "panic", "casualty", and "fractured limb"), but cannot reflect the actual earthquake damage. If the text mining-based analysis is performed without cleaning these noisy tweets, the results will hardly reflect the real earthquake damage situation. Furthermore, the expression style of the social media data typically exhibits characteristics of colloquial language, which did not match the formal keywords extracted from the intensity scale. For example, "A few people who were still indoors or in the high-rise felt it" is used to depict the human perception within the second level of the China Seismic Intensity Scale. However, it is customary for us to adopt a more colloquial and vernacular language style when posting on social media platforms (i.e., "slight", "no feeling", "no shock", "sleep like a log"), rather than employing such scholarly language. Therefore, considering this aspect and establishing a large-language classification model based on social media data would lead to a decrease in classification accuracy and efficiency.

## 2 Overview of related studies

### 2.1 Social media for earthquake analysis

Remote sensing and post-disaster surveying are the most well-known post-earthquake information-acquiring methods (Ahadzadeh & Malek, 2021). However, the remote sensing-based methods have some spatial, temporal, and spectral constraints (Ahadzadeh & Malek, 2021), and the post-disaster surveying-based methods are time-consuming and require extensive labor and resources (Zou et al., 2018; Yu et al., 2019).

In recent years, social media platforms are becoming increasingly popular and closely related to everyone's life. Social media can provide massive amounts of real-time data, which is a new data source and bring new opportunities for post-earthquake management (Yao et al., 2021; Chen et al., 2020). In recent years, many attempts have been made to utilize the contents from social media for earthquake impact analysis (Ogie et al., 2022; Zhang et al., 2019; Li et al., 2023). Most studies on social media content analysis collected data using a keyword search (Zhang et al., 2019). Chen & Ji estimate public demand by leveraging information in sample tweets through keyword filters. Utilize a Bayesian-based method to learn the relationship between Twitter-based demand percentage and survey-based demand percentage, thereby achieving a reliable and rapid estimation of public demand following disasters (Chen & Ji, 2022). Because the contents of the collected microblogs are various, to automatically transform and standardize the unstructured text into a structured form for utilization, fine-grained categorization is explored by many researchers based on NLP methods (Yu et al., 2019; Zhang et al., 2019). For example, Imran et al. classified informative microblogs into five categories, including caution and advice, casualties and damage, donations of money, goods or services, people missing or found, and information sources (Imran et al., 2013). Based on the categories proposed by Imran et al. (2013), additional categories (e.g., infrastructure and resource) are added to their classification schema (Yu et al., 2019). The categories proposed by Imran et al. (Imran et al., 2013) are useful for the response and recovery stage, but may not involve the data released before or after a disaster event (Huang & Xiao, 2015). To consider such information, 47 message categories are defined for preparedness, response, impact, and recovery (Huang & Xiao, 2015). In addition to the categories focusing on different stages, there are also some categories focusing on seismic damage. For example, to extract the seismic intensity, some four or five levels of seismic intensity categories (e.g., no damage, slight damage, moderate damage, and severe damage) are proposed (Yao et al., 2021; Li et al., 2021; Lv et al., 2023). Xing et al. developed a classification standard based on impacts and affected individuals, which primarily concerns social media contents that are interconnected with the relocation and resettlement due to the disaster (Xing et al., 2021). In the meanwhile, Zekkos et al. categorized the tweets into six types, including Automated, Impact, Felt Intensity, Supporting message, Funny and Undetermined (Zekkos et al., 2021).

However, the social media data contain a large amount of noisy microblogs with the keywords of the earthquake but is irrelevant to the real seismic loss of the target earthquake (Kundu et al., 2018; Lin et al., 2022). The above four-level seismic intensity categories can not filter the tweets or retweet that does not describe the real damage (Li et al., 2021), which will make the analysis results differ from the actual situation of the earthquake impact. Furthermore, the expression style of the social media data typically exhibits characteristics of colloquial language, which did not match the formal keywords extracted from the intensity scale. Despite the existing efforts on the classification categories and methods of microblogs, few studies filter weak correlation microblogs before assessment and furthermore analyzing the influence of these weak correlation microblogs, and take the colloquial language style prevalent on social media platforms into account while physical impact assessment. Therefore, a novel category that can consider the correlation between microblogs and earthquake damage should be proposed first.

## 2.2 Text classification methods for social media

The volume of content from microblogs is far beyond the capabilities of manual efforts. Therefore, automated text classification methods are widely used to filter and classify data. Traditional text classification methods include keyword-based methods and machine learning-based (e.g., support vector machine, Naïve Bayes, random forest, logistic regression). Filtering keywords (e.g., damage, injury, and destroy) are used to identify tweets related to earthquake damage (Li et al., 2021). Naïve Bayes-based method is used to classify the microblogs into different categories (Binsaeed et al., 2020). The support vector machine and the latent Dirichlet allocation (LDA) algorithm were combined to classify microblogs (Wang et al., 2016). Ragini & Anand (Ragini & Anand et al., 2016) proposed a classification algorithm with the combination of a support vector machine and Naïve Bayes. While the generalization performance of these traditional methods is not so high, which indicates that these methods may have erroneous results in a new disaster event that is beyond the scope of the designed study (Yu et al., 2019). With the development of deep learning techniques which can achieve a comprehensive understanding of a text to improve performance (Zheng et al., 2022), some deep learning-based text classification models are utilized for social media. For example, Dasari et al. employed a stacking ensemble approach to categorize tweet information. They selected the top five machine learning models with the highest performance as base learners and utilized logistic regression as the meta-learner (Dasari et al., 2023). Specifically, convolutional neural network (CNN)-based models (Yu et al., 2019; Devaraj et al., 2020; Xing et al., 2021) and long short-term memory (LSTM)-based models (Li et al., 2021; Wadud et al., 2022; Arbane et al., 2023) are widely-used deep learning methods. These deep learning models can achieve better performance than the traditional models (Yu et al., 2019; Wadud et al., 2022), while the main drawback of the CNN and LSTM-based models is that they require a huge amount of manpower to prepare enough training data sets (Xu & Cai, 2021). Therefore, pretrained deep learning models (e.g., bidirectional encoder representation from transformers (BERT) (Devlin et al. 2018)) are gaining more research interest in recent years. The parameters of the pretrained models have been adjusted on a large corpus (Devlin et al., 2018) so that the pretrained model can achieve satisfactory performance when the training dataset from a new task is relatively small (Zheng et al., 2022). Furthermore, the BERT model has also been applied in seismic analysis. Lv et al. combined the BERT and the TextCNN model to classify micblogs based on seismic intensity (Lv et al., 2023). Although pretrained models have been used in many applications, in the field of social media, the studies on text classification using the BERT model and how to improve the generalization performance of the BERT model are limited. Therefore, this study employs the BERT model to establish a large-language model for social media data and investigates the influence of different factors on the generalization performance of the BERT model, aiming to develop a more robust earthquake impact analysis system.

## 3 Research objective

Few studies consider filtering weak correlation microblogs before assessment and furthermore analyzing the influence of these weak correlation microblogs. Besides, although pretrained models have been used in many applications, in the field of social media, the studies on how to improve the generalization performance of the BERT model are limited. It also should be noted that, despite the existing efforts have been made to utilize the contents from social media for earthquake impact assessment, there is a notable scarcity of studies that take into account the expression style commonly found in social media data, characterized by colloquial language.

Therefore, this study has the following objectives: 1) Proposes an enhanced approach to address the high noise in microblogs for accurate classification of social media texts. The primary strategy encompasses the introduction of a few categories to classify and filter microblogs based on their relationship to the physical and social impacts of earthquakes. On the basis, the first domain-specific Language Model (LLM) is introduced to achieve accurate classification of social media texts and tackle influential noisy microblogs in earthquake assessment effectively, 2) Proposes an integrated approach to assess both physical and social impacts of earthquakes based on social media texts. By integrating public opinion trend analysis, sentiment analysis, and keyword-based physical impact quantification, it enable effective post-disaster emergency responses to create more resilient cities. Of particular significance is that, this study has taken into account the prevalent colloquial language style observed on social media platforms during the physical impact quantification based on keywords, thereby enhancing the accuracy of earthquake severity rating.

## 4 Methodology

Fig. 1 shows the proposed approach for the accurate classification of social media texts and rapid earthquake impact assessment, which can consider the influence of noisy microblogs. The approach mainly consists of three steps. The first step is data acquisition (Section 4.1). Web crawlers are utilized to acquire the information of microblogs after an earthquake occurs, and then preliminary data cleaning (deduplication) is performed on the raw data. Then, because there are many irrelevant microblogs in the raw data, which will affect the assessment results. Therefore, categories that consider the strength of correlation with the physical and social impacts of earthquakes are proposed. A largescale dataset is also developed for model training. The second step is the development and fine-tuning of the first domain-specific LLM named QuakeBERT for accurate classification and filtering of social media texts (Section 4.2). In addition, the influence of different factors on the generalization performance of the classification model is also explored via several experiments in Section 5. The final step is the earthquake impact assessment (Section 4.3). We used three methods to assess both physical and social impacts of earthquakes, including (1) public opinion trend analysis, (2) sentiment analysis, and (3) keyword-based physical impact analysis. Finally, two earthquake cases with the same magnitude are used for assessment and validation (Section 6). Meanwhile, the python toolkits used to implement the approach are also displayed at the bottom of Fig. 1.

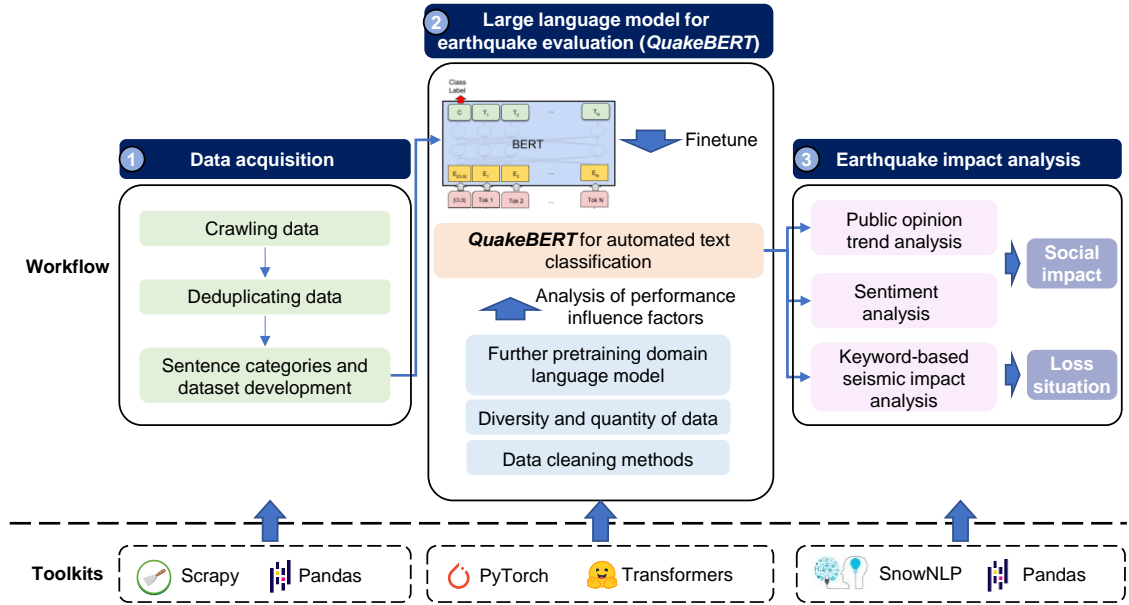

Fig.1. The proposed earthquake impact analysis approach

## 4.1 Data acquisition

### 4.1.1 Data crawling and deduplicating

Based on Scrapy (Kouzis-Loukas, 2016) and Weibo-search (Chen, 2022), we developed a system crawling the microblogs after a specific earthquake. Input the time (e.g., "2008-05-12") and the name of the city (e.g., "Sichuan Wenchuan") where an earthquake occurred into the system, and then the system can automatically obtain and save the information of the microblogs which contain the keywords (i.e., name of the city and "earthquake") and are published after the earthquake occurred. Aiming to help rescue dispatch through earthquake impact analysis, the system exclusively crawls the microblogs within the golden rescue time (i.e., 48 hours) after the earthquake (Macintyre et al., 2006). The crawled information mainly includes textual contents, publication time, user id, and microblog id (each microblog has a unique microblog id). Since the raw data is repetitive, deduplication is performed based on the microblog id.

### 4.1.2 Sentence categories and dataset development

Previous studies indicate that many crawled microblogs are irrelated with the real earthquake impact. If the text mining-based assessment is performed without filtering these noisy tweets, the results will hardly reflect the real earthquake damage situation and lead to ineffective decision making. Additionally, we observe that different types of microblog content tend to describe various aspects of disaster-related information, such as loss description and rescue guidance. Therefore, it is necessary to classify the microblogs before performing an earthquake impact assessment. Through the analysis of contents from earthquake microblogs, the microblogs can be divided into six categories including: "loss description", "education", "notification", "rescue", "related information", and "unrelated". The categories, classification criteria, the strength of correlation and corresponding examples of classification are shown in Table 1.

The category "loss description" includes the microblogs that describe the physical response, damage, and emotion experienced by human beings and the microblogs that describe the physical damage of a specific location. The microblogs in the category "loss description" can reflect the seismic impact on human life and the intensity of the earthquake. Hence, the category "loss description" is strongly correlated with the earthquake's impact.

The category "education" primarily encompasses informative microblogs, frequently associated with science popularization. These microblogs introduce knowledge on self-rescue procedures during an earthquake, which contains the keyword reflecting severe damage (i.e., "panic", "casualty", and "fractured limb"). However, they do not reflect the actual seismic damage.

The category "notification" includes microblogs resembling news reports, typically disseminated by official accounts to report the basic information of an earthquake (e.g., the magnitude, location, time, etc. of an earthquake). This kind of microblog lacks the details of the damage caused by an earthquake.

The category "rescue" includes microblogs that report the news of military or civilian earthquake relief. However, this kind of microblog usually only shows that rescues are being carried out, but lacks details of the rescue, and thus can hardly reflect the real physical damage.

The category "related information" includes the microblogs that talks about the earthquake but are not helpful for physical impact analysis. The content of microblogs in these four categories cannot directly reflect the actual earthquake damage. Nevertheless, their quantity correlates with the severity of an earthquake and reflects the public's level of attention towards it. Therefore, we consider these categories of microblogs to have a weak correlation with the seismic impact.

Besides, the category "unrelated" includes the microblogs that just use the keyword of an earthquake as a hot gimmick, but actually, the actual content has nothing to do with the earthquake. Although the information about an earthquake is mentioned, the main content of the microblog is irrelated to the earthquake, which cannot reflect the impact of the earthquake.

The development of the largescale datasets is illustrated in Section 5.1. And then we develop the first domain-specific LLM based on BERT for accurate classification of microblogs, which is illustrated in Section 4.2.

Table 1 sentence categories

| Category | Relevance | Classification criteria | Example |
|---|---|---|---|
| Loss description | Strong | describe the response, damage, and emotion experienced by human beings or the physical damage of a specific location. | " In the morning, I was woken up by violent shaking while sleeping in a daze. The cupboard also shook and made a loud noise. Fortunately, it stopped after a while. To be honest, I was not afraid at the time, and I didn't even think about running downstairs. " |
| Education | Weak | the knowledge about how to self-rescue during an earthquake | " # Guizhou Bijie 4.4 magnitude earthquake # Earthquake self-rescue knowledge that must be remembered: (1) don't panic when an earthquake occurs, and (2) if the casualty needs a splint to mend the fractured limb, you can use plastic bags and cardboard to make an emergency splint bandage. Please |

| | | | remember these common senses of earthquake prevention and risk avoidance to save yourself at critical moments. " |
| --- | --- | --- | --- |
| Notification | Weak | report the basic information of an earthquake | " # 5.1-magnitude earthquake hits Tangshan # According to the official measurement of the China Earthquake Network, at 6:38 on July 12, Beijing time, a 5.1-magnitude earthquake occurred in Guye District in Tangshan, Hebei Province, which was also felt in Beijing, Tianjin, and other places. " |
| Rescue | Weak | report the news of military or civilian earthquake relief | " After an earthquake of magnitude 7.4 occurred in Maduo county, Guoluo prefecture, Qinghai province at 2:04 a.m. on May 22, the Ministry of Emergency Management and the China earthquake administration sent a rescue team overnight to Maduo county, Qinghai province to guide and assist localities in carrying out emergency rescue and disaster relief work. The fire rescue corps of Qinghai province mobilized five rescue teams from Xining, Hainan, Yushu, Guoluo, and Xunbao to reinforce the disaster area, carry out earthquake rescue and help the citizens resume production and life. " |
| Related information | Weak | talks about the earthquake but is not helpful for physical impact analysis | "# 4.4-magnitude earthquake in Bijie, Guizhou # I hope everyone is safe. The teacher said that Guizhou is the safest city. I didn't expect that there would be another earthquake in Guizhou " |
| Unrelated | No | contain the keywords of an earthquake, but the content is irrelated to the earthquake. | " 3 people from Bijie are on the list! The second quarter of 2022 'Guizhou Good People List' released" |

## 4.2 Development of the domain-specific large language model for earthquake assessment

### 4.2.1 BERT-based text classification method

After establishing the classification category, the first domain-specific LLM based on BERT, QuakeBERT, is developed and fine-tuned for accurate classification and filtering of microblogs. The detailed fine-tuning process and selection of parameters are described in Section 5.3. Here is only a brief introduction to the architecture of the BERT model.

The BERT model mainly consists of three types of layers, including (1) the word embedding layer, (2) the encoding layer, and (3) the classification layer. The word embedding layer consists of three parts, including (1) token embeddings, (2) segmentation embeddings, and (3) position embeddings, so as to realize the conversion of input information into corresponding vector expressions. The encoding layer consists of twelve transformers, which encode all the embeddings from the embedding layer into contextual representations to better understand the contextual semantic information. The classification layer consists of a fully connected layer and a softmax layer, which takes the contextual representation outputted by the encoding layer as input to predict the target categories. The specific details of the BERT model structure can be found in Devlin et al. (2018).

### 4.2.2 Analysis of performance influence factors

The QuakeBERT model is utilized to classify microblogs before further analysis, thus the generalization performance of the model is significant to the subsequent analysis. To investigate the influence of different factors on the performance of the QuakeBERT model, we conducted a series of experiments in Section 5. The main factors include (1) the diversity and volume of the training datasets, (2) domain-specific further pretraining models, and (3) data cleaning methods. These factors were chosen for the following reasons, and the corresponding experiments are illustrated in Section 5.2.2 ~ 5.2.4, respectively.

There are several reasons to investigate how the diversity and volume of the training dataset impact the performance of pretrained models. First, the diversity of the training datasets mainly comes from two aspects, including (1) the distribution of the categories of microblogs varying from different earthquakes, and (2) the characteristics (e.g., geographic location, time) of each microblog varying from different earthquakes. Therefore, a model trained on a single earthquake dataset may not have sufficient generalization capabilities due to variations in data distribution and characteristics. Besides, the volume of the training datasets also has a significant impact. However, it is unclear which of these two factors has a greater impact on the generalization performance of the classification model.

Another factor under investigation is the influence of domain-specific further pretraining language models. In most cases, the general-domain corpus (e.g., news corpus, encyclopedia corpus) is used to pretrain the bert-base-chinese model, whose data distribution is different from the domain corpora targeted to a certain domain (Sun et al., 2019; Zheng et al., 2022). Therefore, directly adopting the bert-base-chinese model is likely to reduce the contribution of domain-specific knowledge embedded in the domain corpora. To investigate the influence of the domain-specific further pretraining models, we pretrain two domain-specific models, bert-earthquake and bert-earthquake-clean, using the corpus with the same distribution as the target domain. The bert-earthquake model is pretrained using the original corpus, while the bert-earthquake-clean model is pretrained using the cleaned corpus, allowing for a concurrent exploration of the impact of data cleaning on pretraining effectiveness, as illustrated in Section 5.2.3.

Lastly, the impact of data-cleaning methods is also analyzed. The performance of the fine-tuning process may be influenced by certain words within the microblog content of datasets, introducing potential confusion during training. To address this, we implemented data-cleaning methods for both datasets, which encompassed the removal of geographic location words, stop words, and topic tags.

It is worth noting that model tends to take shortcuts, relying heavily on location-based information for classification, thereby compromising its overall generalization capability. Therefore, the objective of removing geographic location words is to eliminate location-based information from the text, facilitating earthquake damage analysis independent of varying geographical contexts in each earthquake instance. By removing such information, the model is able to avoid overemphasizing the impact of specific earthquake locations, ensuring a more generalized understanding of earthquake damage. Importantly, it safeguards against the model fixating solely on the limited earthquake locations present in the training set, promoting a more unbiased and robust fine-tuning process.

Similarly, the removal of stop words involves the elimination of commonly used words that do not add any significant meaning to the text.

Finally, the removal of topic tags aims to filter out irrelevant content that is often marked with "#"

symbols. Similar to the aforementioned analysis, the model also tends to take shortcuts by relying on tags for classification. However, many content marked with "#" symbols for many data points does not correspond to the actual content. Specifically, many data in the "Unrelated" category typically only mention earthquake-related keywords in content marked with "#" symbols, but do not provide any further information related to earthquakes, which merely uses the earthquake keyword as a popular gimmick. For exemple, "The air quality in Inner Mongolia is poor. Residents in Chengdu should wear masks when going out and turn on the air purifier at home. #4.4-magnitude earthquake in Bijie, Guizhou#." At the same time, other microblogs strongly or weakly related to the earthquake's impact may contain unrelated information in content marked with "#" symbols, such as the account name reporting the content. For example, "#Southward View of the World# According to the official measurement of the China Earthquake Network, At 07:02 on July 12th, a 2.2-magnitude earthquake occurred in Guye District, Tangshan City, Hebei Province (39.76 degrees north latitude, 118.44 degrees east longitude), with a depth of 15 kilometers. Wishing everyone safety and well-being!" Therefore, to investigate whether the model might be confused by the information in content marked with "#" symbols during training, topic tags are removed during data cleaning.

### 4.3 Integrated method for earthquake impact assessment

After the classification of microblogs, an integrated method integrating (1) public opinion trend analysis, (2) sentiment analysis, and (3) keyword-based physical impact quantification are used to assess both physical and social impacts of earthquakes.

**4.3.1 Public opinion trend analysis**

Public opinion trend analysis uses weakly correlated data and strongly correlated data, which aims to estimate the impact and scope of an earthquake by counting the number of microblogs released in each time period and analyzing the trend of microblogs. This work obtains the releasing time of each microblog in a unified format (YYYY-MM-DD HH:MM) based on Weibo-search. Then, using pandas and matplotlib, the trend of public opinion within 48 hours after the earthquake can be visualized on an hourly basis.

**4.3.2 Sentiment analysis**

Sentiment analysis uses weakly relevant data and strongly relevant data, which is to classify whether the emotion of each microblog is positive or negative. Sentiment analysis of earthquake-related microblogs can estimate the economic losses and social impacts caused by an earthquake. Negative emotions may increase when earthquake damage is severe. Typical positive and negative microblogs are shown in Table 2.

This study employs the sentiment analysis model in snowNLP (a Python library for Chinese NLP tasks) to estimate the sentiment of each microblog. Microblogs with a positive emotion probability greater than 50% are considered positive ones, and vice versa. On this basis, the evolution of the number and proportion of two types of microblogs can be calculated and visualized.

Table 2. Typical microblogs with two different sentiments

| Sentiment | Text |
| --- | --- |

| | | |
|---|---|
| Positive | On April 6, a class of a middle school in Xingwen County, Yibin, Sichuan was reading early when an earthquake warning sounded. The students quickly hunched over their heads and squatted to avoid it, and then began to evacuate in an orderly manner. The students did not panic, and there were no casualties. |
| Negative | In Xingwen County, the Qishupo section and the Xinhuachang section of the provincial highway (S444) collapsed, causing road interruptions. At present, traffic control has been implemented on the above-mentioned road sections, and vehicles and pedestrians are prohibited from passing through! |

**4.3.3 Keyword-based physical impact quantification**

Keyword-based physical impact quantification uses strongly relevant data, which can reflect the actual loss of earthquakes. First, we established a table of physical impact assessment based on intensity keywords, as shown in Table 3. Table 3 shows the different levels of earthquake impact and the corresponding keywords. According to the seismic intensity scale of China (GB/T 17742–2020) and the previous studies (Yao et al., 2021; Li et al., 2021), the impact of earthquakes can be divided into four levels. Subsequently, the keywords corresponding to each level are preliminarily determined according to the descriptions of the conditions at each level in the seismic intensity scale of China (GB/T 17742–2020). By analyzing the microblogs, we found that the keywords extracted from the intensity scale were formal, which did not match the colloquial language habit of contents in the microblogs. This mismatch between the informal linguistic patterns and the formal keywords extracted from the intensity scale has a detrimental impact on the accuracy of assessment. Therefore, we have modified and supplemented the keywords to be more suitable for language habits and peoples' perceptions. Finally, we obtained the intensity keyword table, as shown in Table 3.

Then the microblogs are classified into four levels utilizing the keyword matching method based on the keywords in Table 3. If a sentence contains keywords in a certain disaster level, this sentence is classified into this level. In some cases, the words belonging to the lower level only have a few more qualifiers than the higher level (e.g., " slight cracks " in the second level and " cracks " in the third level). If high-level words are matched first, then low-level words will also be matched, so that the low-level words will be misjudged as high-level words. So if a sentence contains multiple keywords in different levels at the same time, the classification priority is first level > second level > third level > fourth level. If a sentence contains no keyword, then we filter this sentence because it provided no information related to the earthquake damage.

Table 3. Keywords for different disaster level

| Disaster level | Human feeling | House damage | Other phenomena | Casualties | Lifeline condition |
|---|---|---|---|---|---|
| First level | no feeling, no shock, sleep like a log, didn't know there was an earthquake, can't wake up ( in Chinese: 没有感觉, 没 | slight, good ( in Chinese: 轻微, 良好 ) | slight vibration (in Chinese: 微动 ) | no one was injured (in Chinese: 没有人员受伤) | normal, no damage found, flat, no cracks ( in Chinese: 正常, 未发现受损 |
| | | 震感, 睡死, 不知道地震, 叫不醒) | | | 情况, 平整, 无裂缝) |
| Second level | feel, standing unsteadily, fleeing outside in a panic, shaking, feeling the shock, wake up, waking up, running downstairs, kicking, not daring to move (in Chinese: 有感, 站立不稳, 惊逃室外, 摇, 有震感, 摇醒, 震醒, 往楼下跑, 踢, 不敢动) | noise, dust falling, fine cracks, eaves tiles falling, bricks falling, slight damage, basically intact, shaking, trembling, slight crack, falling off (in Chinese: 响, 灰土掉落, 细裂缝, 檐瓦掉落, 掉砖, 轻微破坏, 基本完好, 晃动, 颤动, 轻微开裂, 脱落,) | swinging, ringing, shaking, overturning, moving, slight cracks, sandblasting, watering (in Chinese: 摆动, 响, 摇动, 翻倒, 移动, 轻度裂缝, 喷沙, 冒水) | trapped, diverted, minor abrasions, minor injuries (in Chinese: 被困, 转移, 轻微擦伤, 轻微伤) | restricted, blocked, damaged, slight cracks, no major impact (in Chinese: 管制, 阻塞, 受损, 细微裂缝, 未造成较大影响) |
| Third level | bumping, falling, difficulty walking, obvious prolonged, dizzy, avoiding danger, unsteady, shaking vigorously, staggering, unbalanced (in Chinese: 颠簸, 摔倒, 行走困难, 明显, 晕, 避险, 站不稳, 使劲摇, 跟跟跄跄, 失衡) | moderately damaged, cracks (in Chinese: 中等破坏, 裂, 开裂) | drop, crack, moderate damage, misalignment, tilt, shock, fall (in Chinese: 掉落, 裂缝, 中等破坏, 错动, 倾斜, 震落, 倒) | buried, wounded, fractured, crushed, trapped, buried under pressure, concussion, badly wounded (in Chinese: 埋, 受伤, 骨折, 被压, 被困, 被埋压, 脑震荡, 重伤) | prohibited, closed, impassable, rockfall, damaged, torn, rolling stones, backed up, tripped, interrupted, cracks (in Chinese: 禁止, 封闭, 无法通行, 落石, 损坏, 撕裂, 滚石, 退服, 跳闸, 中断, 开裂) |
| Fourth level | falling away from the original place, the feeling of being thrown up, violent, fierce, strong, serious (in Chinese: 摔离原地, 有抛起感, 剧烈, 猛, 凶, 强烈) | severely damage, destroy, collapse (in Chinese: 严重破坏, 毁坏, 坍塌) | collapse, severe damage, rupture, destruction, toppled, landslide, severe (in Chinese: 倒塌, 严重破坏, 断裂, 破坏, 倒毁, 滑坡, 严重) | dead, distress, misfortune, danger, death (in Chinese: 死亡, 遇难, 不幸, 生命危险, 死) | break, landslide, subsidence, deformation, heave, collapse, shock collapse, collapse, collapsed, rolling stone (in Chinese: 断裂, 滑坡, 塌方, 变形, 隆起, 坍塌, 沉陷, 震垮, 垮塌, 塌了, 滚石) |

## 5 Experiments on the performance of text classification models

### 5.1 Dataset development

#### 5.1.1 Training dataset development

  Based on scrapy (Wang et al., 2012) and Weibo-search (Chen, 2022), we crawled the disaster-related microblogs of nine different earthquakes in different locations. Subsequently, we filtered the repetitive microblogs in the raw data and then manually labeled them to form the training dataset. A total number of 6268 disaster-related microblogs were labeled. The locations of the crawled data are shown in Table 4, and the category distribution of the labeled data is shown in Fig. 2.

Table 4 Locations of the crawled data

| Earthquake information | Number of the labeled microblogs |
| --- | --- |
| 2020-07-12 5.1 magnitude earthquake in Tangshan, Hebei | 3170 |
| 2019-06-17 5.1 magnitude earthquake in Yibin, Sichuan | 1106 |
| 2017-03-27 5.1 magnitude earthquake in Dali, Yunan | 187 |
| 2018-10-31 5.1 magnitude earthquake in Liangshan, Sichuan | 331 |
| 2022-06-20 4.4 magnitude earthquake in Bijie, Guizhou | 434 |
| 2021-11-17 5.0 magnitude earthquake in Yancheng, Jiangsu | 50 |
| 2022-03-17 5.1 magnitude earthquake in Zhangye, Gansu | 134 |
| 2021-03-19 6.1 magnitude earthquake in Naqu, Xizang | 231 |
| 2021-05-22 7.4 magnitude earthquake in Guoluo, Qinghai | 625 |
| Sum | 6268 |

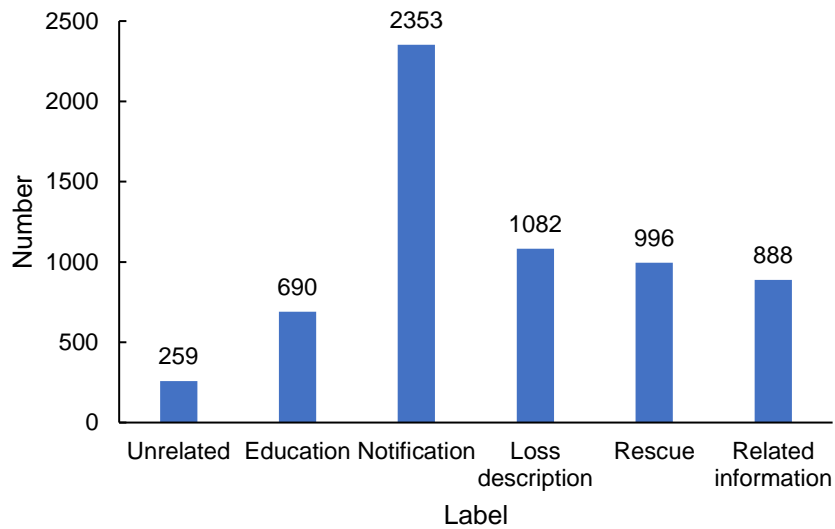

Fig. 2 Category distribution of the training dataset

Fig. 2 shows that the amount of data in the "unrelated" category is relatively small, which means the training dataset is imbalanced. The imbalanced dataset is not conducive to model training. Therefore, we collected 751 other irrelevant data from different earthquakes to supplement our training dataset. After supplementing, the training dataset has a total of 7019 labeled microblogs, and the data distribution is shown in Fig. 3.

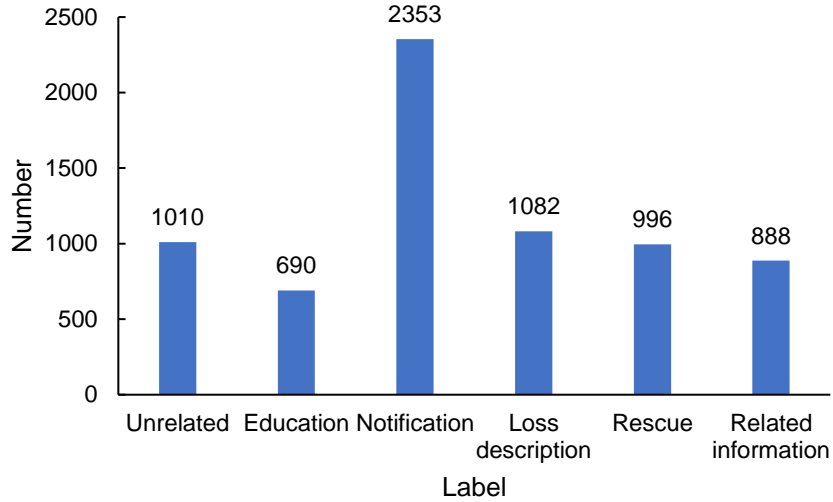

Fig. 3 Category distribution of the training dataset after supplementing

**5.1.2 Test dataset development**

The test dataset is used to assess the generalization ability of the well-fine-tuned model. Because the characteristics and distribution of data from different earthquakes are different, we choose the various earthquakes that are not contained in the training datasets. Then we labeled a total of 1014 microblogs as the test dataset, to assess the generalization performance of different models. The locations of the crawled data are shown in Table 5, and the category distribution of the labeled test dataset is shown in Fig. 4.

Table 5 Distribution of test datasets

| Earthquake information | Number of the labeled microblogs |
|---|---|
| 2020-09-13 Akesu, Xinjiang | 29 |
| 2020-10-06 Ali, Xizang | 22 |
| 2020-10-21 Mianyang, Sichuan | 74 |
| 2020-12-12 Wulumuqi, Xinjiang | 79 |
| 2021-01-04 Leshan, Sichuan | 136 |
| 2021-01-23 Zhaotong, Yunnan | 70 |
| 2021-05-13 Baoshan, Yunnan | 92 |
| 2021-07-14 Aba, Sichuan | 160 |
| 2021-07-23 Luzhou, Sichuan | 61 |
| 2021-08-26 Jiuquan, Gansu | 64 |
| 2022-01-08 Haibei, Qinghai | 306 |
| Sum | 1014 |

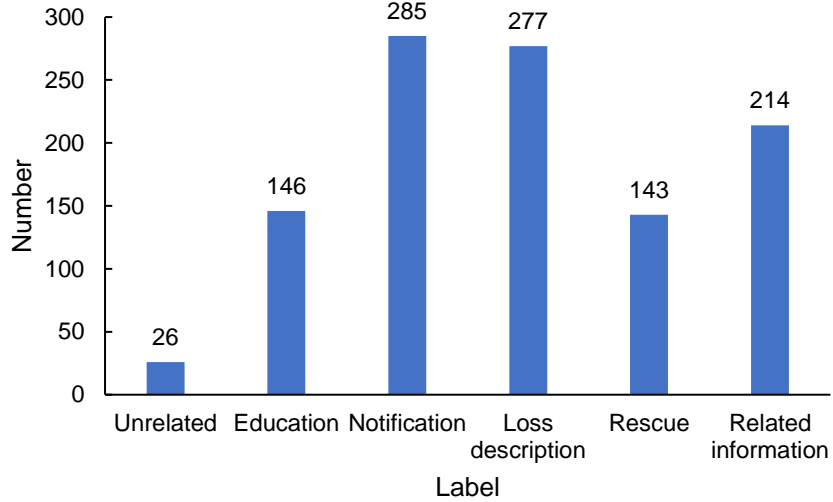

Fig. 4 Category distribution of the test dataset

## 5.2 Experiment settings

The analyses of (1) the diversity and volume of the training datasets, (2) domain-specific further pretraining models, and (3) data cleaning methods are illustrated in Section 5.2.2 ~ Section 5.2.4, respectively. To measure the generalization performance of the models, we first constructed a test dataset (Section 5.1.2) that did not appear in the training datasets, and then the macro average F1 score (Zheng et al., 2022; Vani et al., 2019; Zhou et al., 2022) is used to evaluate models, as illustrated in Section 5.2.1.

### 5.2.1 Evaluation performance metrics

The predictions by the models are compared with the gold standard to evaluate the performance. Because the datasets are highly unbalanced, the macro average F1 score (macro F1) is selected.

First, we calculate the precision (P), recall (R), and F1 score (F1) for each category as follows:

$$P = N_{correct}/N_{labeled} \tag{1}$$

$$R = N_{correct}/N_{true} \tag{2}$$

$$F_1 = 2PR/(P + R) \tag{3}$$

where $N_{\{correct, labeled, true\}}$ denotes the number of {model correctly labeled, model labeled, true} elements for each category.

Then, we calculate the macro average F1 score to represent the overall performance (m denotes the number of categories):

$$\text{Macro } F_1 = \left(\sum_{i}^{m} F_{1,i}\right)/m \tag{4}$$

### 5.2.2 Diversity and volume of the training datasets

We conducted three experiments to investigate the impact of diversity and volume of training datasets. As for the diversity of the dataset, we conducted experiment 1. The model trained on the dataset containing microblogs from one earthquake ("DS-VL" (short for the dataset with low diversity and large volume)) and the model trained on the dataset containing microblogs from multiple earthquakes ("DL-VL" (short for the dataset with large diversity and large volume)) are compared. Note that the two datasets have the same amount of samples, as shown in Table 4. As for the volume of datasets, we conducted the second and third experiment. In experiment 2, we compare two models trained on the datasets with the same distribution and different quantities, which are "DL-VL" and "DL-VEL" (short for the dataset with large diversity and even larger volume) in Table 4. Besides, in order to eliminate the influence of increased data diversity, we conducted experiment 3, which two models were traind on the datasets "DS-VS" (short for the dataset with low diversity and low volume) and "DS-VL".The construction process of datasets is shown in Section 5.1.

Table 4 Experiments on the diversity and volume of the training datasets

| Group name | Training dataset | Model |
|---|---|---|
| DS-VS | Tangshan (1585) | bert-base-chinese |
| DS-VL | Tangshan (3170) | bert-base-chinese |
| DL-VL | various earthquakes (3170) | bert-base-chinese |
| DL-VEL | various earthquakes (7091) | bert-base-chinese |

Note: DS-VS is short for the dataset with low diversity and low volume. DS-VL represents the dataset with low diversity and large volume. DL-VL is the dataset with large diversity and large volume. DL-VEL is the dataset with large diversity and even larger volume. Tangshan (3170) is short for 3170 microblogs from the 2020-07-12 earthquake in Tangshan, Hebei; Tangshan (1585) is obtained by randomly selecting half of the data from each class in the Tangshan (3170); various earthquakes (3170) is short for 3170 microblogs from various earthquakes; various earthquakes (7091) is short for 7091 microblogs from various earthquakes.

**5.2.3 Further pretrained domain-specific language models**

We first crawled 120,000 Sina microblogs as the original corpus. The original corpus contains some noisy texts including geographic location texts, stop words, and topic tags. Implementing data cleaning may improve the performance of the further pretrained models, as illustrated in Section 4.2.2. However, since the original corpus contains a large number of earthquakes of various types, the impact of noise and irrelevant information may be minimal. Additionally, the original corpus, without performed data cleaning, maintains a stronger integrity, which is more conducive to the model understanding the semantic information of microblog content. Therefore, to investigate whether data cleaning improves model effectiveness, we conducted data cleaning to obtain the cleaned corpus.

Then the mask language model (Devlin et al., 2018) is utilized to pretrain two domain-specific models, named bert-earthquake and bert-earthquake-clean. The bert-earthquake model is pretrained using the original corpus, and the bert-earthquake-clean model is pretrained using the cleaned corpus. The models are pretrained with a learning rate of $5 \times 10^5$ and an epoch of 5, while also used a 15% probability of masking each token (Devlin et al., 2018; Zheng et al., 2022).

On this basis, to investigate the influence of pretraining on the performance of text classification, we conducted twelve experiments. The corresponding training datasets and pretraining models are shown in Table 5.

Table 5 Experiments on domain-specific further pretraining models

| Model | Tangshan (1585) | Tangshan (3170) | various earthquakes (3170) | various earthquakes (7091) |
|---|---|---|---|---|
| bert-base-chinese | Group 1 | Group 2 | Group 3 | Group 4 |
| bert-earthquake | Group 5 | Group 6 | Group 7 | Group 8 |
| bert-earthquake-clean | Group 9 | Group 10 | Group 11 | Group 12 |

Note: Tangshan (3170) is short for 3170 microblogs from the 2020-07-12 earthquake in Tangshan, Hebei; Tangshan (1585) is obtained by randomly selecting half of the data from each class in the Tangshan (3170); various earthquakes (3170) is short for 3170 microblogs from various earthquakes; various earthquakes (7091) is short for 7091 microblogs from various earthquakes.

**5.2.4 Data cleaning methods**

A series of experiments were conducted to explore the impact of data cleaning methods on the model performance, as follows. The "bert-base-chinese" model is used in this experiment. As illustrated in Section 4.2.2, data cleaning methods include stop-word removal, geographic location text removal, and topic tag removal. In the experiment, the data cleaning methods are performed on the training and test datasets. Ablation experiments were conducted to investigate the influence of the three data cleaning strategies. The experimental settings are shown in Table 6. In these experiments, we added one more data cleaning strategy at a time to the training dataset. First, we performed stop-word removal (i.e., "S training" in Table 6). Then stop-word removal and removal of geographic location text (i.e., "SG training" in Table 6) were preformed. Finally, we employed all three data cleaning strategies simultaneously (i.e., "SGT training" in Table 6). Furthermore, we analyzed the effect of data cleaning on the test dataset (i.e., "SGT test" in Table 6). Intuitively, cleaning the test dataset is likely to reduce the semantic complexity of the sentences to be classified, thereby improving the classification performance.

Table 6 Experiments on data cleaning methods

| | Tangshan (1585) | | Tangshan (3170) | | various earthquakes (3170) | | various earthquakes (7091) | |
|---|---|---|---|---|---|---|---|---|
| | O test | SGT test | O test | SGT test | O test | SGT test | O test | SGT test |
| O training | Group 1 | Group 2 | Group 3 | Group 4 | Group 5 | Group 6 | Group 7 | Group 8 |
| S training | Group 9 | Group 10 | Group 11 | Group 12 | Group 13 | Group 14 | Group 15 | Group 16 |
| SG training | Group 17 | Group 18 | Group 19 | Group 20 | Group 21 | Group 22 | Group 23 | Group 24 |
| SGT training | Group 25 | Group 26 | Group 27 | Group 28 | Group 29 | Group 30 | Group 31 | Group 32 |

Note: Tangshan (3170) is short for 3170 microblogs from the 2020-07-12 earthquake in Tangshan, Hebei; Tangshan (1585) is obtained by randomly selecting half of the data from each class in the Tangshan (3170); various earthquakes (3170) is short for 3170 microblogs from various earthquakes; various earthquakes (7091) is short for 7091 microblogs from various earthquakes. O is short for original; S is short for utilizing stop-word removal; SG is short for utilizing stop-word removal and geographic location text removal; SGT is short for utilizing stop-word removal, geographic location text removal, and topic tag removal.

## 5.3 Experiment results

To find the model with the best performance, the training dataset is randomly split into the training dataset and the validation dataset at a 0.8 : 0.2 ratio, where the training dataset is used to train and update the BERT model, and the validation dataset is used to choose the best combination of the hyperparameters and the best model. The test dataset is used to test the best model performance as generalization performance.

Furthermore, the impact of different learning rates ($7 \times 10^{-5}$, $5 \times 10^{-5}$, $3 \times 10^{-5}$, and $1 \times 10^{-5}$) and batch sizes (8, 16, and 32) were considered by grid search. The optimizer is Adam, and other parameters are defaulted except for the initial learning rate. We found out that the model starts to overfit on training dataset at around the 10 epoch, so we set the max epoch to 30, which far exceeds the epoch that the optimal model requires. Besides, to reduce training time, we add an early stop mechanism. If the model performance on the validation dataset does not improve for 3 epochs, the training process will be terminated.

### 5.3.1 Influence of diversity and volume of the training dataset on model performance

The experiment's results are listed in Table 9. Comparing the results of "DS-VL" and "DL-VL", it can be found that the performance of the model is significantly improved if the diversity of datasets increases. The macro average F1 score increases by 26.37%. Comparing the results of "DS-VS" and "DS-VL", as well as "DL-VL" and "DL-VEL", it can be found that increasing data volume significantly improves model performance when data diversity is low, with the macro average F1 score increasing by 15.4%. However, the improvement is relatively small when data diversity is high, showing a macro average F1 score increase of 0.7%. Generally, the impact of variables on model performance diminishes with an increase in data volume. However, in scenarios with a large data volume, enhancing data diversity has a more pronounced effect on model performance improvement (26.37%) compared to increasing data volume in situations with a lower dataset (15.4%). Therefore, we can conclude that data diversity is more critical than data volume for enhancing the performance of the microblog classification model.

Table 9 Experiment results on the diversity and volume of the training dataset

| Group name | Training datasets | Macro F1 |
| --- | --- | --- |
| DS-VS | Tangshan (1585) | 0.4186 |
| DS-VL | Tangshan (3170) | 0.5726 |
| DL-VL | various earthquakes (3170) | 0.8363 |
| DL-VEL | various earthquakes (7190) | **0.8433** |

Note: DS-VS is short for the dataset with low diversity and low volume. DS-VL represents the dataset with low diversity and large volume. DL-VL is the dataset with large diversity and large volume. DL-VEL is the dataset with large diversity and even larger volume. Tangshan (3170) is short for 3170 microblogs from the 2020-07-12 earthquake in Tangshan, Hebei; Tangshan (1585) is obtained by randomly selecting half of the data from each class in the Tangshan (3170); various earthquakes (3170) is short for 3170 microblogs from various earthquakes; various earthquakes (7091) is short for 7091 microblogs from various earthquakes.

### 5.3.2 Performance of the further pretrained domain language models

The experiment results are listed in Table 10. Comparing the results of three pretrained models trained using the Tangshan (1585) and the Tangshan (3170), the bert-earthquake-clean model achieves the best performance, and the bert-earthquake model performs better than the original model. It can be found that when the diversity of the training dataset is small, further pretraining models on the domain corpus can improve the performance of the models. However, comparing the results of three pretrained models trained using the various earthquakes (3170) and the various earthquakes (7091), the bert-base-chinese model achieves the best performance. The F1 scores for each category of three pretrained models trained using the various earthquakes (7091) are listed in Table 11. It can be observed that the standard deviation of the F1 scores predicted by the bert-base-chinese model across different categories is around 0.05, indicating a relatively strong consistency and stability compared to other models. This model demonstrates robust performance across different categories of microblogs, ensuring reliable predictions under various circumstances. Furthermore, the results show that the bert-base-chinese model has the highest prediction accuracy for the "loss description" category, which is strongly correlated with the earthquake's impact. This aspect facilitates subsequent analyses of the physical and social impacts of earthquakes. Consequently, it is evident that when the diversity of the training dataset is large, further pretraining methods may not significantly enhance model performance.

Table 10 Experiment results on further pretraining BERT models

| Model | Tangshan (1585) | Tangshan (3170) | various earthquakes (3170) | various earthquakes (7091) |
|---|---|---|---|---|
| bert-base-chinese | 0.4186 | 0.5726 | **0.8363** | **0.8433** |
| bert-earthquake | 0.4272 | 0.5792 | 0.7334 | 0.7966 |
| bert-earthquake-clean | **0.4335** | **0.6103** | 0.7931 | 0.8253 |

Note: Tangshan (3170) is short for 3170 microblogs from the 2020-07-12 earthquake in Tangshan, Hebei; Tangshan (1585) is obtained by randomly selecting half of the data from each class in the Tangshan (3170); various earthquakes (3170) is short for 3170 microblogs from various earthquakes; various earthquakes (7091) is short for 7091 microblogs from various earthquakes.

Table 11 The F1 scores for each category on further pretraining BERT models

| Category | **bert-base-chinese** | bert-earthquake | bert-earthquake-clean |
|---|---|---|---|
| Unrelated | **0.9057** | 0.7636 | 0.7931 |
| Education | **0.8897** | 0.8689 | 0.9143 |
| Notification | **0.8390** | 0.8187 | 0.8431 |
| Loss description | **0.8212** | 0.8111 | 0.7950 |
| Rescue | **0.8563** | 0.8179 | 0.8263 |
| Related information | **0.7480** | 0.6995 | 0.7802 |
| **Macro average** | **0.8433** | 0.7966 | 0.8253 |
| **Standard Deviation** | **0.0562** | 0.0581 | 0.0495 |

### 5.3.3 Influence of data cleaning on model performance

The experiment's results are listed in Table 12. Comparing the performance of the models trained using the Tangshan (1585) and the Tangshan (3170), the model trained on the "SGT training" dataset achieves the best performance on the cleaned test dataset (i.e., "SGT test" in Table 12). Besides, the performance of all models on the "SGT test" dataset is better than that on the original test dataset, which may be due to the reduced semantic complexity of the "SGT test" dataset. When utilizing the Tangshan (1585) dataset, the top-performing model achieved an overall best macro average F1 score of 0.4922, exhibiting an improvement of 11.71% over the original result of 0.3751. Similarly, with the Tangshan (3170) dataset, the overall best macro average F1 score reached 0.6538, showcasing an 8.12% enhancement compared to the original result of 0.5726. It can be found that when the diversity of the training dataset is small, the performance of the model can be improved by cleaning the test dataset.

However, comparing the performance of the models trained using the various earthquakes (3170) and the various earthquakes (7091), the original model achieves the best performance. Besides, the performance of all models on the "SGT test" dataset is lower than that on the original test dataset (i.e., "O test"). The reason may be that the "SGT test" dataset lacks some features that the model uses for classification. So, when the diversity of the training dataset is large, data-cleaning of the training dataset and test dataset can not improve the performance of the model.

Table 12 Experiment results on data cleaning methods

|  | Tangshan (1585) | | Tangshan (3170) | | various earthquakes (3170) | | various earthquakes (7091) | |
| --- | --- | --- | --- | --- | --- | --- | --- | --- |
|  | O test | SGT test | O test | SGT test | O test | SGT test | O test | SGT test |
| O training | 0.4186 | 0.4725 | 0.5726 | 0.6423 | **0.8363** | 0.7384 | **0.8433** | 0.7187 |
| S training | 0.3980 | 0.4707 | 0.5735 | 0.6146 | 0.8226 | 0.7266 | 0.8278 | 0.7278 |
| SG training | 0.3806 | 0.4858 | 0.56 | 0.6385 | 0.7734 | 0.6943 | 0.8191 | 0.7131 |
| SGT training | 0.3751 | **0.4922** | 0.5773 | **0.6538** | 0.765 | 0.7646 | 0.7615 | 0.7723 |

Note: Tangshan (3170) is short for 3170 microblogs from the 2020-07-12 earthquake in Tangshan, Hebei; Tangshan (1585) is obtained by randomly selecting half of the data from each class in the Tangshan (3170); various earthquakes (3170) is short for 3170 microblogs from various earthquakes; various earthquakes (7091) is short for 7091 microblogs from various earthquakes. O is short for original; S is short for utilizing stop-word removal; SG is short for utilizing stop-word removal and geographic location text removal; SGT is short for utilizing stop-word removal, geographic location text removal, and topic tag removal.

### 5.3.4 Performance comparison with other models

After the above analysis, the best model is the "bert-base-chinese" model trained on the dataset named "various earthquakes (7091)", which is named QuakeBERT. The QuakeBERT is used for the model comparison and further physical impact analysis.

We further compared our models with three widely-used deep learning-based text classification methods, including (1) TextCNN (Chen, 2015), (2) TextRNN with attention mechanism (TextRNN-Att for short) (Cai et al., 2018), and (3) TextRCNN (Lai et al., 2015). The initial parameters of the word embedding layers of these models are pretrained on the Chinese Wikipedia corpora (Li et al., 2018) via skip-gram models (Mikolov et al., 2013). Then, the labeled data of 7091 microblogs from various earthquakes are used to train the above models. 100 epochs, far more than the best model required, and a padding size of 256, the same as the BERT-based model, are selected. The effects of different learning rates and batch sizes are considered via grid searches with batch sizes of 64, 32, and 16, and with learning rates of 0.002, 0.001, 0.0005, 0.00025, and 0.0001. The models with the best performance on the validation dataset are used for evaluation on the test dataset.

The performance of different text classification models is shown in Fig.5. The QuakeBERT outperformed the other deep learning models, which improved the macro F1 score of 23.46%. The performance on the test dataset shows that the QuakeBERT model has strong generalization performance.

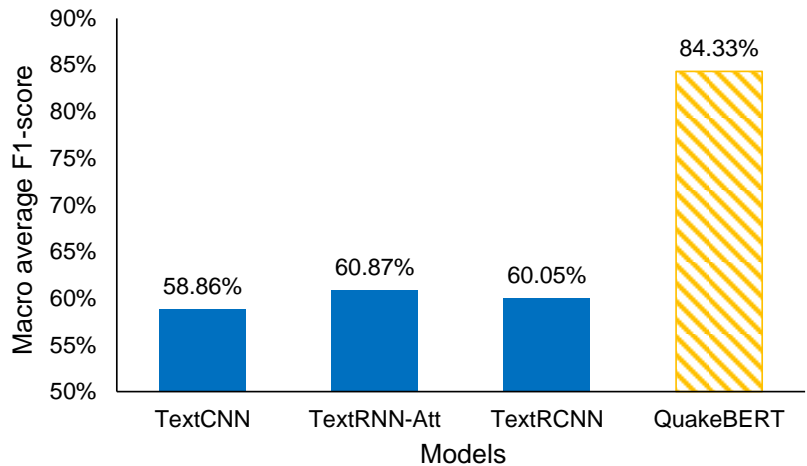

Fig. 5 Performance of different models

# 6 Case study

In this section, we validate our classification method by analyzing the impact of two earthquakes: the Xingwen earthquake in Sichuan Province, China, and the Guye earthquake in Hebei Province, China, both of which have the same magnitude.

**6.1 Earthquake information and data collection**

The Xingwen earthquake occurred at 7:50 on April 6, 2022, with the epicenter located at 28.22 N and 105.03 W, with a magnitude of 5.1, and a depth of 10 km. Its maximum intensity was 6 degrees, and the 6-degree zone included seven townships, with an area of approximately 379 km$^2$. Approximately 750,000 people were affected, with 719 houses damaged, 1,045 residents urgently transferred, 2 provincial roads interrupted, and 5 village roads damaged.

The Guye earthquake occurred at 6:38 on July 12, 2020, with the epicenter located at 39.78 N and 118.44 W, also with a magnitude of 5.1, and a depth of 10 km. Its maximum intensity was 5 degrees, covering four districts, including Guye, Kaiping, Luanzhou, and Qian'an, with an area of approximately

437 km² and an affected population of about 2.36 million. The earthquake effects only included minor cracks in individual old houses.

After data collection and deduplication, the data volume for the Xingwen earthquake was 2087, and the data volume for the Guye earthquake was 5156. Table 13 summarizes the key information about these two earthquakes.

Table 13 Earthquake information

|  | Xingwen Earthquake | Guye Earthquake |
|---|---|---|
| Magnitude | 5.1 | 5.1 |
| Depth | 10 km | 10 km |
| Damage | 719 houses damaged, 1,045 residents urgently transferred, 2 provincial roads interrupted, 5 village roads damaged | minor cracks in individual old houses |
| Volume of data | 2087 | 5156 |

## 6.2 Results of data analysis and impact assessment

First, the well-trained QuakeBERT model with the best generalization performance in Section 4 is used to classify the collected microblogs to determine their correlation with the physical and social impacts of earthquakes. The results are presented in Table 14, which reveals that over 50 microblogs are not related to the target earthquake keywords. It is essential to filter out these unrelated microblogs for accurate analysis since they can negatively impact the results. Subsequently, we filtered out the unrelated microblogs using the QuakeBERT model and proceeded with the remaining ones for further analysis. Public opinion trend analysis and sentiment analysis use weakly correlated data and strongly correlated data. Keyword-based physical impact analysis uses strongly relevant data, which can reflect the actual loss of earthquakes.

Table 14 sentence categories

| Category | Relevance | Guye | Xingwen |
|---|---|---|---|
| Loss description | Strong | 864 | 289 |
| Education | Weak | 466 | 518 |
| Notification | Weak | 2784 | 283 |
| Rescue | Weak | 750 | 438 |
| Related information | Weak | 240 | 556 |
| Unrelated | No | 52 | 3 |
| Sum |  | 5156 | 2087 |

### 6.2.1 Public opinion trend analysis

Fig. 6 depicts the temporal evolution of public opinion in the 48 hours following the occurrence of the two earthquakes. Both earthquakes experienced a surge in public attention within the first hour after the event, followed by a rapid decline. It is noteworthy that the evolution tends to end near 48 hours.

However, it should be emphasized that the peak volume of public opinion only indicates the level of attention the earthquake event received, and is not necessarily reflective of the actual damage caused. Taking this study as an example, the Xingwen earthquake resulted in relatively high losses. Yet, its peak volume of public opinion was significantly lower compared to that of the Guye earthquake. This is due to the fact that the Xingwen earthquake occurred in an area with a population of only 750,000, which is significantly lower than the population of approximately 2.36 million in the area affected by the Guye earthquake. Furthermore, the Guye earthquake was also felt in nearby large cities such as Beijing and Tianjin, contributing to a higher volume of online discussions.

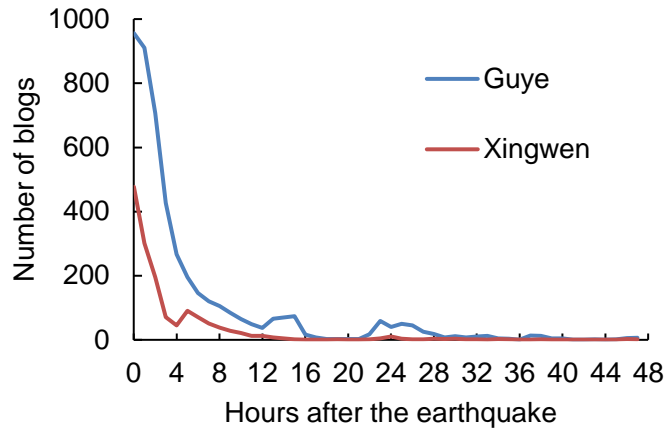

Fig. 6 Public opinion trend of the two earthquakes

**6.2.2 Sentiment analysis**

Fig. 7 (a) & 7 (b) depict the trends of positive and negative microblogs within 48 hours after the Guye and Xingwen earthquakes, respectively. The proportion of negative emotions in microblogs after the Xingwen earthquake was significantly higher than that of the Guye earthquake, which is consistent with the loss situation discussed in Section 6.1. The content of microblogs in the early period after an earthquake mainly comprises spontaneous discussions by people in the vicinity of the epicenter and surrounding areas affected by the earthquake. Therefore, a higher proportion of negative microblogs within 2 hours after an earthquake indicates a more powerful earthquake with more severe losses. These results align with previous studies, which suggest that sentiment may be even more sensitive to damage (Kryvasheyeu et al., 2016) and that citizens tend to express negative sentiment on social media during a disaster (Wu & Cui, 2018). Furthermore, as disaster relief and rescue efforts progress, the sentiment of microblogs tends to become more positive over time.

Thus, monitoring the emotional proportion of microblogs in real-time after an earthquake can provide a reference for early disaster relief decisions and help estimate the damage caused by the earthquake.

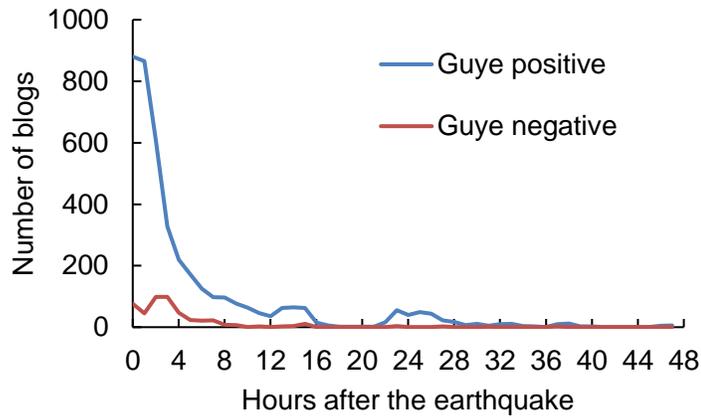

(a) Guye

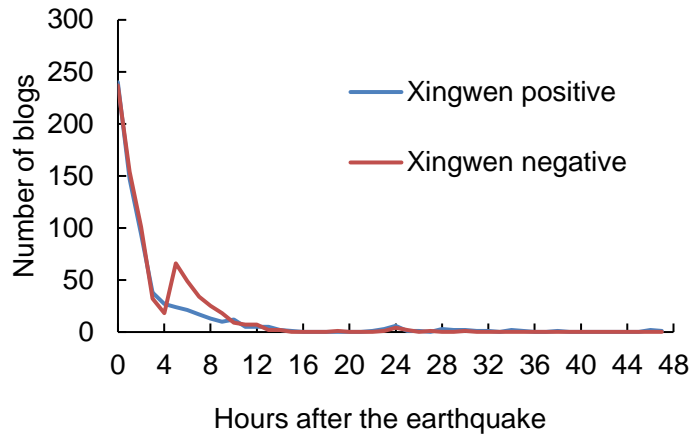

(b) Xingwen

Fig. 7. Sentiment trend of the two earthquakes (a) Guye (b) Xingwen

**6.2.3 Keyword-based physical impact quantification**

Using the keyword-based physical impact analysis method described in Section 4.3.3, the disaster levels of each microblog of the two earthquakes are assessed. To demonstrate the effectiveness of the proposed classification method, we compared the analysis results using the original unclassified data with the strongly correlated data (i.e., the data in the "loss description" category) classified by the QuakeBERT model. The number of microblogs with different disaster levels is shown in Fig. 8. From Fig. 8, we can observe the following phenomena. (1) the data has a low signal-to-noise ratio, and noise data has a significant impact on determining the severity of earthquake losses. (2) The distribution of microblogs with different loss levels before and after classification is inconsistent, indicating that noise data will affect the assessment of earthquake loss. (3) From the analysis of the two earthquakes, it was observed that the signal-to-noise ratio at the second and third levels was relatively lower, which may cause misleading results in the analysis process. For example, this can result in underestimating the damage for earthquakes with severe damage and overestimating the damage for earthquakes with low damage.

Fig. 9(a) shows that, before classification, the proportion of microblogs in the fourth level (the most severe level) of the Guye earthquake exceeded that of the Xingwen earthquake, suggesting that the impacts of the Guye earthquake were more severe than those of the Xingwen earthquake. However, the assessment results did not align with the actual loss situation, as discussed in Section 6.1. This could be attributed to the presence of microblogs with weak or no correlation in the unclassified dataset. Such microblogs contain many keywords listed in Table 3, but they do not reflect the specific earthquake loss situation, thereby affecting the assessment results. On the other hand, only the strongly correlated microblogs were retained for analysis after classification. Thus, Fig. 9(b) shows that after classification, the proportion of microblogs in the fourth level for the Guye earthquake was lower than that for the Xingwen earthquake, indicating that the impacts of the Guye earthquake were weaker than those of the Xingwen earthquake. The assessment results were consistent with the actual loss situation.

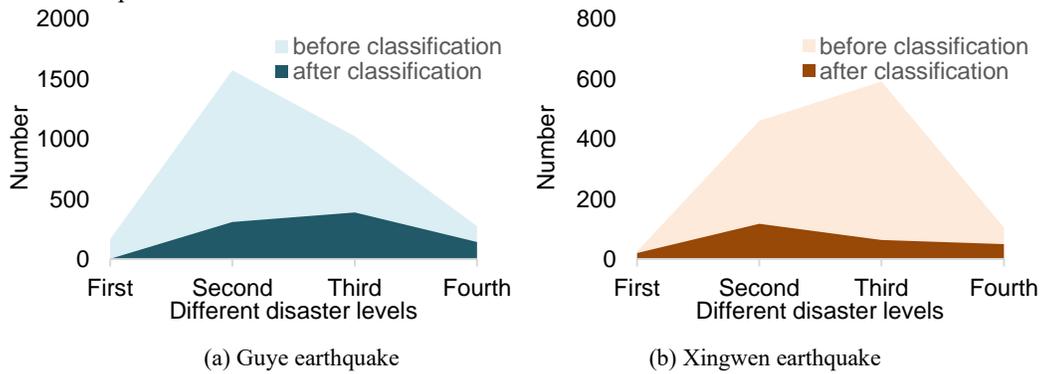

Fig. 8. Number of microblogs with different disaster levels (a) Guye earthquake (b) Xingwen earthquake

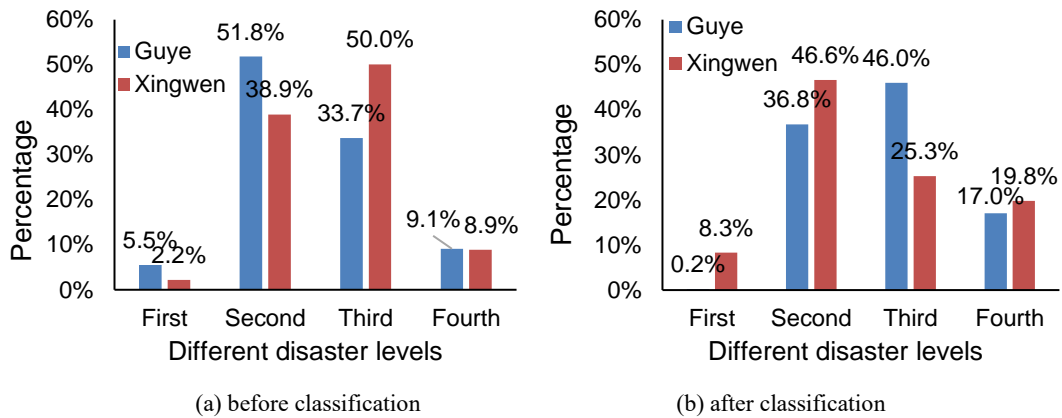

Fig. 9. Percentage of microblogs with different disaster levels (a) before classification (b) after classification

# 7 Discussion

**7.1 Comparison with previous research**

Despite the existing efforts on the the classification categories and earthquake analysis of microblogs, many of them lack comprehensive consideration, as shown in Table 15.

Table 15 Literature summary

| Literature | large-scale model | physical impacts | social impacts | colloquial representation | noisy microblogs |
| --- | --- | --- | --- | --- | --- |
| Yao et al., 2021 | × | √ | × | √ | × |
| Li et al., 2021 | × | √ | × | × | × |
| Xing et al., 2021 | √ | √ | × | × | × |
| Lv et al., 2023 | × | √ | × | × | × |
| Dasari et al., 2023 | √ | × | × | × | × |
| This study | √ | √ | √ | √ | √ |

This work proposed an enhanced approach to accurate classification of social media texts for rapid earthquake impact assessment. Compared to previous research, this work contributes to the body of knowledge on four main levels:

(1) To reduce the influence of noisy microblogs, this work proposes the first domain-specific large language model (LLM) for accurate classification and filtering of microblogs related to earthquakes. First, sentence categories that consider the relevance of microblogs are established, and the largescale training and testing datasets are established. Then based on the experiments, a BERT-based model for microblog classification (QuakeBERT) with high generalization performance is obtained. Subsequently, after classification, public opinion trend analysis, sentiment analysis, and keyword-based physical impact analysis were carried out to assess the physical and social impacts of earthquakes. Finally, two earthquakes of Xingwen and Guye with the same focal depth and magnitude are used for comparative analysis. The analysis results show that noisy microblogs have a significant influence on the keyword-based physical impact analysis method, and will induce errors in the physical assessment. The proposed text classification method can reduce the effect of noisy microblogs and improve the accuracy of keyword-based physical impact analysis.

(2) To improve the generalization performance of the QuakeBERT model, this work analyzes several influential factors, including the diversity and volume of the training dataset, further pretraining models, and data cleaning methods. The results show that the diversity and volume of the training dataset have the greatest impact on the generalization performance. In scenarios with a large data volume, enhancing data diversity has a more pronounced effect on model performance improvement (26.37%) compared to increasing data volume in situations with a lower dataset (15.4%). When the diversity and volume of training data are small, data cleaning and further pretraining models are beneficial to improve the generalization performance. Besides, the QuakeBERT model outperformed the other deep learning models, which improved the macro F1 score of 23.46%.

(3) To achieve rapid physical impact quantification, this study established a keyword database related to rapid physical loss based on the seismic intensity scale of China. Then the keyword database was modified and supplemented to be more suitable for language habits and peoples' perceptions of social media. Utilizing the keyword database, keyword-based physical impact analysis can be performed.

(4) To illustrate the advantages and disadvantages of different analysis methods for assessing the physical and social impacts of earthquakes, two earthquakes of Xingwen and Guye with the same magnitude and focal depth are used for comparative analysis. The results can reach the following conclusions. 1) The magnitude of the earthquake is the same, but the physical losses caused by the earthquake are different. 2) The results of public opinion trend analysis are related to the degree of people's attention but have little correlation with the actual earthquake loss. 3) The proportion of negative emotions in the sentiment analysis is related to the actual physical loss. The more proportion of negative emotions is, the more severe the physical damage caused by an earthquake. 4) The keyword-based physical impact analysis will be easily affected by the noisy microblogs. After removing the noise microblogs, the analysis results of keyword-based physical impact analysis have a high correlation with the actual physical loss.

**7.2 Theoretical contributions**

In order to address the deficiencies in existing literature regarding the accurate classification of microblogs and the rapid assessment of earthquake impacts, we conducted this study, making significant contributions in the following aspects:

First, we introduce a few categories to classify and filter microblogs based on their relationship to the physical and social impacts of earthquakes, considers noisy microblogs that fail to reflect the actual extent of earthquake damage. This categorization scheme serves as a quantitative criterion for assessing the physical impacts of earthquakes.

Second, to expend, we propose the first domain-specific Language Model (LLM) to achieve accurate classification of social media texts and tackle influential noisy microblogs in earthquake assessment effectively. This approach significantly enhances classification accuracy and facilitates the acquisition of intricate and varied characteristics exhibited in earthquake-related microblogss. It specifically considers the presence of noisy microblogs, and other relevant features, thereby enhancing the overall performance of the classification process.

Finally, we contribute to an integrated approach to assess both physical and social impacts of earthquakes based on social media texts, considers the usage of colloquial language. By integrating public opinion trend analysis, sentiment analysis, and keyword-based physical impact quantification, it enable effective post-disaster emergency responses to create more resilient cities.

**7.3 Practical contributions**

In this study, we provide a more precise representation of the actual situation regarding earthquake impacts, enabling more effective post-disaster emergency responses to create more resilient cities, primarily manifested in the following aspects:

First, it is imperative to consider the substantial impact of noisy microblogs on earthquake impact assessment. The enhanced approach porposed in this study enables rapid and accurate classification of earthquake-related social media texts, effectively addresses these issue, and facilitates rapid quantification of physical impacts.

Second, the governments' time-sensitive responses rely on timely and accurate information about the earthquake situations to make effective response decisions and improve management strategies. The integrated approach proposed in this study allows for assessment of both physical and social impacts of earthquakes. Compared to previous researchs, this study considers the prevalent colloquial language style observed on social media platforms and provides quantitative evaluation of physical impacts and qualitative evaluation of social impacts, offering a more comprehensive and rapid assessment. This approach provides decision-makers and managers with more effective decision support.

**7.4 Limitations and future works**

Limitations to this study need to be addressed in future work:

(1) The amount of microblogs that can be obtained by using the web crawler is limited. In future work, it is recommended to cooperate with social media companies (e.g., Twitter and Sina Weibo) to obtain more data.

(2) In addition, due to privacy considerations, the geo-location of tweets is not available unless users actively elect. This is the main obstacle to quantitative and accurate analysis of earthquake damage in specific locations. Therefore, in the future, multiple data source integrating methods should be considered for rapid earthquake damage analysis. For example, mobile phone signaling data can be integrated to provide effective location information (Xing et al., 2021).

# 8 Conclusion

Earthquakes have a profound impact on human societies, and timely access to information about the scope and extent of disasters is essential for post-disaster decision-making and emergency relief operations. However, traditional post-disaster damage assessment methods are time-consuming and require substantial labor and resources. Social media data have become more and more valuable for the development of resilient and sustainable cities, which can help assess damage at a lower cost. However, the social media data are usually overwhelmed with plenty of noisy information, which will influence the analysis results. This study proposed the first domain-specific LLM for accurate classification and filtering of microblogs to enhance earthquake impact assessment. Specifically, first, this work establishes sentence categories that can consider the relevance of microblogs with the physical and social impacts of earthquakes and a largescale datasets. Then, a BERT-based classification model with high generalization performance (QuakeBERT) is trained to filter and classify noisy microblogs. Subsequently, a keyword database related to rapid physical loss assessment was established for the keyword-based physical impact analysis. Finally, the study conducts a comparative analysis of two earthquakes of Xingwen and Guye with the same magnitude and focal depth. The analysis results show the following conclusions:

(1) Noisy microblogs have a significant influence on the keyword-based physical impact analysis method and will induce errors in the physical assessment.

(2) The proposed classification method based on a large-language model can reduce the effect of noisy microblogs and improve the accuracy of keyword-based physical impact quantification.

In addition, this work systematically analyzes several influential factors (i.e., the diversity and volume of the training dataset, further pretraining BERT model, and data cleaning methods) of the generalization performance of the QuakeBERT model. The results show the following conclusions:

(1) The data diversity and data volume have the greatest influence on the generalization performance of the model.

(2) When the diversity and volume of training data are small, data cleaning and further pre-training models are beneficial to improve the generalization performance of the model.

(3) The QuakeBERT model outperformed the other deep learning models, which improved the macro average F1 score of 23.46%. This shows that the QuakeBERT model has better generalization performance and is more suitable for filtering irrelevant microblogs.

To summarize, the proposed approach, datasets, and large-language models can provide valuable insights for the development of resilient and sustainable cities with rapid impact assessment of earthquakes based on social media texts.

## Acknowledgment

The authors are grateful for the financial support received from the National Natural Science Foundation of China (No. 52238011, 72091512) and the Tencent Foundation through the XPLORER PRIZE.